\documentclass[runningheads]{llncs}
\usepackage[colorlinks,
            linkcolor=blue,       
            anchorcolor=blue,  
            citecolor=blue,        
            ]{hyperref}
\usepackage[misc]{ifsym}

\usepackage{graphicx}
\usepackage{subfigure}
\usepackage{amsfonts}
\usepackage{booktabs}
\usepackage{latexsym}
\usepackage{multirow}
\usepackage{array}
\usepackage{cleveref}
\usepackage{algorithm}
\usepackage{algorithmicx}
\usepackage{algpseudocode}
\usepackage{color}

\begin{document}
\title{Inductive Spatial Temporal Prediction Under Data Drift with Informative Graph Neural Network}
%
%
\author{Jialun Zheng\Letter, Divya Saxena, Jiannong Cao, Hanchen Yang, Penghui Ruan}
\institute{Department of Computing, The Hong Kong Polytechnic University, Hong Kong, China\\
\email{22069255r@connect.polyu.hk}, \email{\{divsaxen,csjcao\}@comp.polyu.edu.hk}, \email{\{hanchen.yang,penghui.ruan\}@connect.polyu.hk}}
\authorrunning{J. Zheng et al.}
\titlerunning{INF-GNN}
%

%
\maketitle              
\pagenumbering{gobble}
\begin{abstract}
Inductive spatial temporal prediction can generalize historical data to predict unseen data, crucial for highly dynamic scenarios (e.g., traffic systems, stock markets). However, external events (e.g., urban structural growth, market crash) and emerging new entities (e.g., locations, stocks) can undermine prediction accuracy by inducing data drift over time. Most existing studies extract invariant patterns to counter data drift but ignore pattern diversity, exhibiting poor generalization to unseen entities. To address this issue, we design an Informative Graph Neural Network (INF-GNN) to distill diversified invariant patterns and improve prediction accuracy under data drift. Firstly, we build an informative subgraph with a uniquely designed metric, Relation Importance (RI), that can effectively select stable entities and distinct spatial relationships. This subgraph further generalizes new entities' data via neighbors merging. Secondly, we propose an informative temporal memory buffer to help the model emphasize valuable timestamps extracted using influence functions within time intervals. This memory buffer allows INF-GNN to discern influential temporal patterns. Finally, RI loss optimization is designed for pattern consolidation. Extensive experiments on real-world dataset under substantial data drift demonstrate that INF-GNN significantly outperforms existing alternatives.

\end{abstract}
\section{Introduction}
Inductive spatial temporal prediction places high demand on generalization of unseen data and is indispensable for highly dynamic application scenarios such as earth science \cite{yang2023higrn}, urban transportation \cite{xu2020ge,cui2019traffic,deng2021graph} and public health \cite{liu2021prototypical}. Existing methods that employ spatial temporal kriging based on generation models \cite{tang2020joint,xu2020ge} or matrix completion approaches \cite{deng2021graph} suffer from data drift, which naturally occurs in evolving spatial temporal data \cite{liu2021prototypical,zhang2022dynamic}.



\begin{figure}[t]
\centering
\subfigure{\includegraphics[width=12cm]{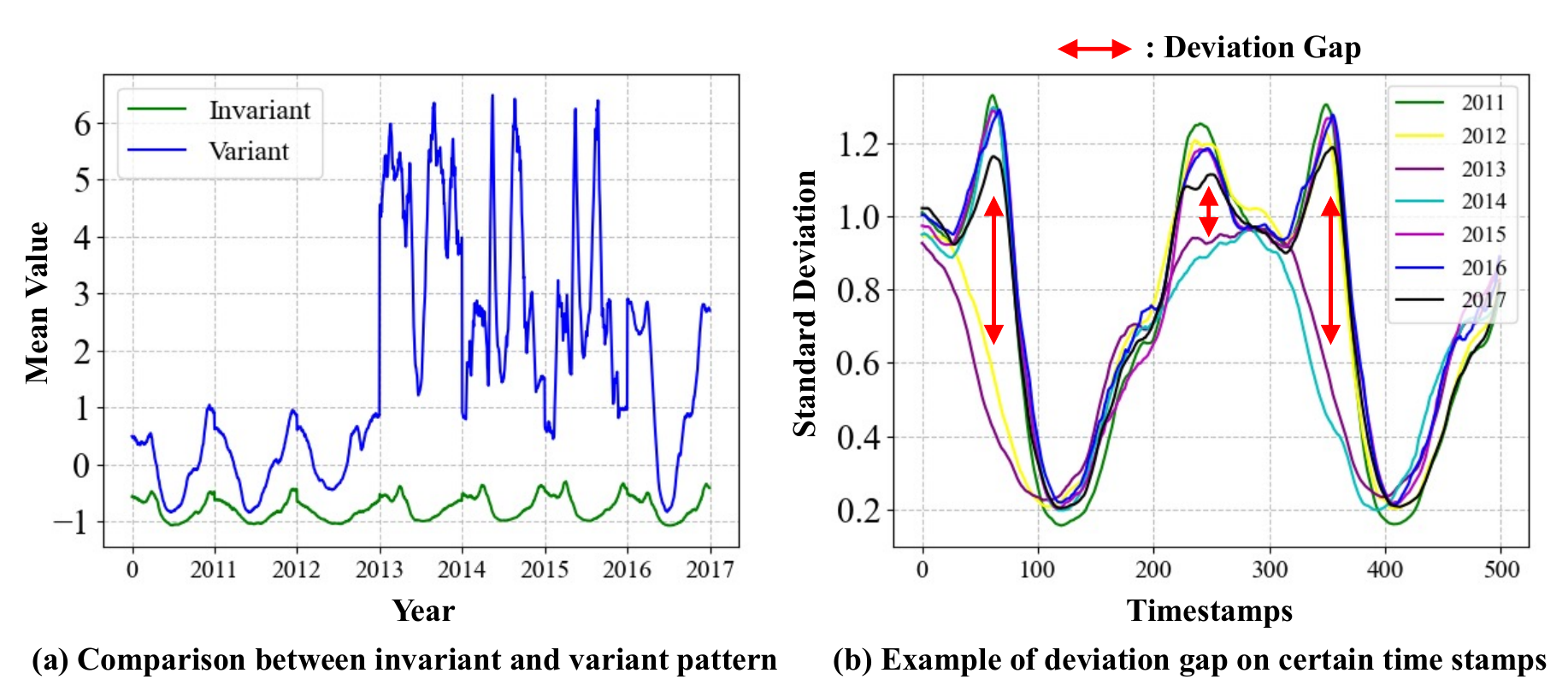}}
\caption{Motivating experiments. (a) Entities with invariant patterns will have distribution remain stable over time. (b) Certain timestamps will have considerable deviation gaps across different time intervals.} \label{fig1}
\vspace{-0.5cm}
\end{figure}

To address this issue, it is important to identify invariant spatial temporal patterns that remain stable under data drift, as they can aid model in performing inductive spatial temporal prediction.
Several online learning models such as \cite{chentrafficstream,wang2023pattern} attempt to extract invariant temporal patterns through building subset of all entities, which only retain entities with stable distribution over time as shown in Fig.~\ref{fig1}(a). However, there exist two major limitations: (1) They only consider temporally invariant patterns while ignoring spatially informative patterns. (2) They treat temporal patterns equivalently within time intervals without focusing on influential timestamps. These limitations hinder the extraction of informative patterns.

Firstly, extracting temporal invariant patterns alone is insufficient, as spatially informative patterns are also necessary for the model to learn diverse entity distributions and achieve better generalizability to unseen entities.
For example, given two entities with one of them has distribution that is noticeably different from its neighboring entities, while another is highly similar to its neighbors. The former can be more spatially informative with higher distribution deviation among nodes and cover much more information. Nevertheless, these two types of entities were regarded equally in existing methods, which indicates that these models are not spatially informative and contain redundancy.

Furthermore, existing works measure the stability of entities by comparing the divergence of their distributions between different time intervals, thereby capturing temporal patterns in general. However, they fail to emphasize influential temporal patterns within specific time intervals.
As shown in Fig.~\ref{fig1}(b), there exist huge deviation gaps on specific timestamps among each time interval. These timestamps need to be treated with more focus as they contain valuable temporal patterns that can assist model in improving generalizability.

To overcome two limitations mentioned above, we design an Informative Graph Neural Network (INF-GNN) to capture informative and invariant spatial temporal patterns for inductive spatial temporal prediction under data drift. 
Our proposed method first develops a Relation Importance (RI) metric based on temporal invariant patterns to select nodes with informative spatial relationships.
Based on these selected nodes, an Informative Subgraph is constructed to simulate new entities.
Then, we uniquely build an informative temporal memory buffer that records valuable timestamps selected by the influence function. These selected timestamps can help model emphasize influential temporal patterns during a time interval. 
Finally, we proposed RI loss optimization to consolidate learned patterns.
To evaluate our proposed framework and show state-of-art performance, We apply our model to perform a prediction task on real-world long-term traffic flow dataset that possess evolving spatial temporal dependencies and entity numbers. This paper provides following contributions:
\begin{itemize}
    \item[$\bullet$] We propose an Informative Graph Neural Network (INF-GNN) to handle inductive spatial temporal prediction under data drift by capturing invariant and informative spatial temporal patterns.
    \item[$\bullet$] We design RI metric to select entities with spatial informative and temporal invariant patterns to construct informative subgraph for simulating new entities. We establish an informative temporal memory buffer to help model emphasize influential timestamps within time intervals. We adopt RI loss optimization to consolidate learned knowledge.
    \item[$\bullet$] Experiments show our method achieves the best performance in the prediction of new entities and existing entities under data drift among all baselines.
\end{itemize}

\section{Related Work}

\subsection{Inductive Spatial Temporal Learning}
Existing inductive spatial temporal learning methods can be roughly divided into two categories:  (1) Spatial temporal kriging \cite{cressie2015statistics,zheng2023increase,wu2021inductive,appleby2020kriging}. (2) Continual spatial temporal prediction \cite{chentrafficstream,wang2023pattern}. The former was designed for interpolating missing values and can be formulated as a matrix completion problem \cite{deng2021graph}. Ge-gan \cite{xu2020ge} adapts generative adversarial networks (GAN) to generate data for unseen entities. Others \cite{wu2021inductive,appleby2020kriging} choose to rely on the inductive power of graph neural networks (GNN). There also exist studies utilizing attention mechanism to fuse spatial temporal patterns for interpolating \cite{zheng2023increase}. However, the kirging method commonly assumes the unseen entities to be under same distribution with historical data and the attention based or GAN based methods often lack interpretability.

The second category is largely based on continual learning (CL) \cite{kirkpatrick2017overcoming,wang2020streaming} due to its promising ability in adapting to new tasks, which can be formulated as predicting new entities in inductive spatial temporal learning. TrafficStream \cite{chentrafficstream} is the first to propose continual learning on spatial temporal learning, which combines the CL framework with GNN and recurrent neural network (RNN) \cite{andreoletti2019network} to continually capture evolving spatial temporal patterns. Notwithstanding, they focus on temporal stable patterns while neglecting spatial informative patterns.

\subsection{Spatial Temporal Data Drift}
Data drift is a innate feature of spatial temporal data \cite{pareja2020evolvegcn,jin2021community,wang2022causal,liu2021coinet,liu2021graph}, especially in areas updated rapidly such as traffic networks and financial market. Existing works usually consider data drift from the perspective of temporal such as the Adarnn \cite{du2021adarnn} uses adaptive RNN models to handle data drift and RevIN \cite{kim2021reversible} adapts normalization to counter data drift. However, they ignore spatial data drift and are limited to statistic normalization \cite{passalis2019deep} that may lose information. 

Borrowing ideas from casual inference \cite{pearl2016causal}, DIDA \cite{zhang2022dynamic} proposed a dynamic graph neural network to capture invariant spatial temporal dependencies that are sufficient for prediction under spatial temporal data drift. Though the results were promising, they are limited to fixed entities numbers and the generalization ability of model is still under explored. Fortunately, our work considers spatially informative patterns in informative subgraph as well as influential temporal patterns stored in memory buffer. These informative invariant patterns enable INF-GNN to achieve high prediction accuracy under spatial temporal data drift triggered by external events and expanding number of entities.

\section{Preliminary}
In this section, We formulate spatial temporal data as dynamic graph and refer to it as dynamic spatial temporal graph. Then we define spatial temporal data drift and inductive spatial temporal prediction.
\begin{definition}
(Dynamic Spatial Temporal Graph). Dynamic spatial temporal graph can be denoted as $G = \{G_1, G_2, .... G_T\}$ with $T$ being number of time intervals, and each interval consists $M$ number of timestamps. Specifically for each time interval $t \in \{1,2,...,T\}$, we have $G_t \in \{G_1, G_2, .... G_T\}$. We further have $G_t = (V_t, E_t, A_t)$, $V_t$ is nodes set consisted of entities, $E_t$ is edges set consisted of spatial relations and $A_t \in \mathbb{R}^{N_t \times N_t}$ is the adjacency matrix in time interval $t$ where $N_t = \vert V_t\vert$ is the number of nodes. During certain time interval $t$, nodes will record spatial temporal data that can be represented as $X^{V_t} \in \mathbb{R}^{N_t \times D \times M}$ where $D$ is the dimension of data.
\end{definition}
\begin{definition}
(Spatial Temporal Data Drift). Graph structure will change when time interval shifts from $t$ to $t+1$ as $G_t \neq G_{t+1}$. Despite the graph structure, spatial temporal data recorded by similar nodes in different time intervals will also be different. Under data drift, we are able to find set $V^{\prime} \subseteq V_t \cap V_{t+1}$, such that $p_t(X^{V^{\prime}}) \neq p_{t+1}(X^{V^{\prime}})$, where $p_t(X^{V^{\prime}})$ and $p_{t+1}(X^{V^{\prime}})$ are nodes set $V$'s recorded spatial temporal data distribution in time $t$ and $t+1$ respectively.
\end{definition}
\begin{definition}
(Problem Definition: Inductive Spatial Temporal Prediction). Given $M$ timestamps spatial temporal data from previous time interval $t$, we aim to predict $K$ timestamps data in next time interval $t+1$ using function $\Psi$:
\end{definition}

\begin{equation}
\Psi(X^{V_t}_1, X^{V_t}_2, ..., X^{V_t}_M) = \{X^{V_{t+1}}_1, X^{V_{t+1}}_2, ..., X^{V_{t+1}}_K\}.
\end{equation}

\section{Methodology}
\begin{figure}[t]
\includegraphics[width=12cm]{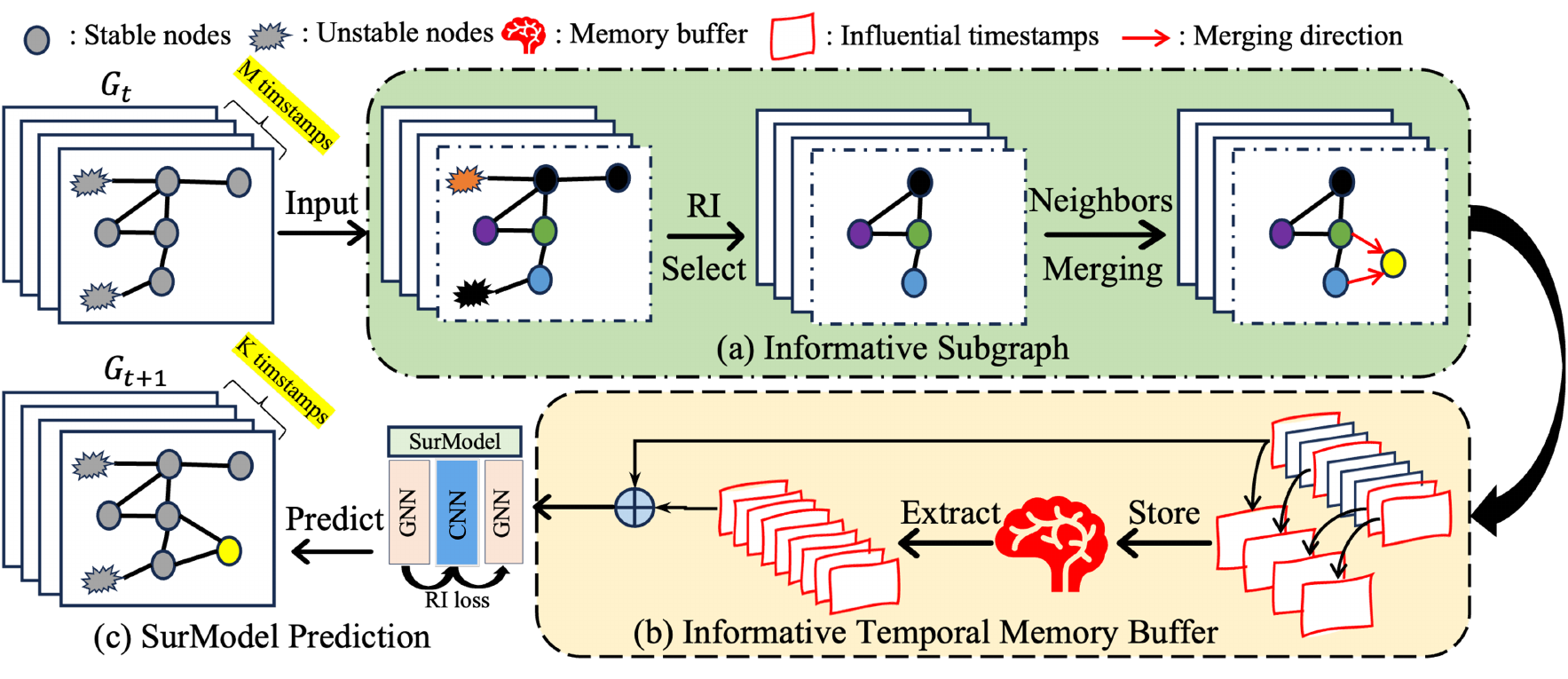}
\caption{General framework of INF-GNN. (a) Using RI metric to select nodes that are stable and have little mutual information with their neighbors to construct informative subgraph, which are further used for simulation of new entities. (b) Selecting informative timestamps by influence function to jointly train with all timestamps. (c) A simple surrogate spatial temporal predicting model is adapted with RI loss optimization to make predictions. } \label{fig2}
\end{figure}

Fig.~\ref{fig2} shows the framework of INF-GNN, which consists of an informative subgraph construction procedure, followed by an informative temporal memory buffer selection and then SurModel with RI loss optimization. Firstly, nodes in informative subgraph are selected by our RI metric, which are further used for generalizing simulation of new entities for subsequent timestamps by employing a neighbors merging technique. Then, informative temporal memory buffer stores timestamps selected by influence functions. These influential timestamps will be further extracted and jointly trained with training data. Finally, we employ a simple surrogate model (SurModel) with RI loss optimization guided parameter updating to make predictions.
\subsection{Surrogate Spatial Temporal Predicting Model}\label{sec4.1}
To ensure our proposed framework is effective without additional benefits from non-trivial neural network design such as attention mechanism, we utilize a simple surrogate spatial temporal predicting mode that is composed of two GNNs and one CNN added between.\\
\indent Given the input to the $l$-th GNN layer in time interval $t$ as $H_t^l \in \mathbb{R}^{N_t \times D^l}$ where $D^l$ is the node feature dimension, the graph convolution operation will change it to $l+1$ layer representation as follows:
\begin{equation}
H_{t}^{l+1} = \sigma  (A_tH_t^lW_1^l + H_t^lW_2^l),
\end{equation}
where $W_2^l, W_1^l \in \mathbb{R}^{ D^l\times D^{l+1}}$ are learnable parameter matrix and $\sigma$ is the activation function. 

To extract temporal patterns, the embedding is then input to a 1D CNN layer and then followed by a GNN layer and a fully connected layer to map $M$ timestamps in time interval $t$ to $K$ timestamps in time interval $t+1$. 

\subsection{Informative Subgraph}\label{sec4.2}
Invariant patterns with informativeness are vital for countering data drift and improving generalizability. To select stable nodes with these patterns to form informative subgraph, we propose a novel metric called Relation Important (RI) with strong interpretability by jointly quantifying stability and informativeness. Stability is indicated by stronger mutual information between a node's features at time $t$ and $t-1$, representing consistent patterns over time. Informativeness refers to weaker mutual information between a node's features and those of its neighbors at both $t$ and $t-1$, denoting independence from neighboring features representations. Based on above motivation and intuition, the RI should be formulated in fractional form as follows:

\begin{equation}
RI(v) = \sum_{u\in \mathcal{N}(v)}\frac{JSD(P_t(u)||P_{t-1}(u))JSD(P_t(v)||P_{t-1}(v))}{JSD(P_t(u)||P_t(v))JSD(P_{t-1}(u)||P_{t-1}(v))}.
\end{equation}
where $\mathcal{N}(v)$ indicate the k-hop neighbor around node $v$. $P_t(u), P_t(v)$ refer to the distribution of $u$ and $v$'s feature in time $t$ and $P_{t-1}(u), P_{t-1}(v)$ refer to the $t-1$'s distribution of nodes $v$ and $u$. JSD refers to Jensen–Shannon divergence, which is a measurement of mutual information that can be calculated as:

\begin{equation}
JSD(P(X)||P(Y)) = \frac{1}{2} D(P(X)||\overline P) + \frac{1}{2}D(P(Y)||\overline P),
\end{equation}
\begin{equation}
D(P(X)||\overline P) = \textstyle\sum_{x \in X}P(x)\log(P(x)/\overline P),
\end{equation}
\begin{equation}
D(P(Y)||\overline P) = \textstyle\sum_{y \in Y}P(Y)\log(P(y)/\overline P),
\end{equation}
\begin{equation}
\overline P = (P(X)+P(Y))/2,
\end{equation}

\noindent where $P(X)$ and $P(Y)$ are two distributions and higher JSD indicates weaker mutual information between two distributions.

The numerator of RI captures nodes' stability by calculating the mutual information between their features over successive timestamps. Lower values in the numerator indicate stronger mutual information, or more consistent patterns over time and is a sign of stability. On the other hand, the denominator of RI measures node's informativeness through the mutual information between its features and those of its neighboring nodes, at both timestamps $t$ and $t-1$. Higher values in the denominator indicate weaker mutual information, pointing to distinction from neighbors and is a hallmark of informativeness.

In this way, lower RI scores are achieved by nodes exhibiting both a lower numerator (higher stability) and higher denominator (greater informativeness). Therefore, RI can explicitly select nodes that display the desired properties of being stable in their patterns while also differing informatively from neighboring nodes and the computation procedure is also explainable and traceable.

We assign each node with RI scores and select those with lowest RI scores to build informative subgraph that has following definition: 

\begin{definition}
(Informative Subgraph). For dynamic spatial temporal network $G_{t-1}$ and $G_t$, there exist a induced subgraph as $G_{if} = (V_{if}, E_{if}, A_{if})$ where $V_{if}\subset V_t\bigcap V_{t-1}, E_{if}\subset E_t\bigcap E_{t-1}$, such that for any induced subgraph $G_{s} = (V_{s}, E_{s}, A_{s})$ where $V_{s}\subset V_t\bigcap V_{t-1}, E_{s}\subset E_t\bigcap E_{t-1}$. If we have $|V_{s}| = N_{s} = N_{if} = |V_{if}|$ and following condition is met:
\begin{equation}
\sum_{v^{\prime}\in V_{if}} RI(v^{\prime})\le \sum_{v\in V_{s}}RI(v).
\end{equation}
then we call induced subgraph $G_{if}$ the informative subgraph
\end{definition}

\noindent We can build our informative subgraph on top of nodes with lowest RI scores, given the fixed $N_{if}$ number. Since these nodes possess lowest RI scores among all nodes in nodes intersection $V_t\bigcap V_{t-1}$, their RI scores sum should also be the lowest to meet the requirement of informative subgraph.

\subsection{Informative Temporal Memory Buffer}\label{sec4.3}
We build informative temporal memory buffer to help model emphasize valuable timestamps, whose recorded data exhibits severe fluctuation between time intervals and encompass informative temporal patterns.

To select such valuable timestamps for the memory buffer, we introduce influence function \cite{koh2017understanding}, which is developed for quantifying the effect of perturbing individual training points on learned model parameters. In this way, timestamps yielding data that significantly deviates from typical patterns will exert greater influence over the learned representations and have higher influence score, thus being recorded by memory buffer and for model to review frequently.

The core idea of influence function is adding a small perturbation to the training batch $\mathcal{B}$ as:
\begin{equation}
\hat{\theta}_{\mathcal{E},\mathcal{B}} = argmin \mathcal{L}(\mathcal{B},\theta) + \mathcal{E}^\mathrm{T}\mathcal{L}(\mathcal{B},\theta),
\label{equ:theta_eb}
\end{equation}
where $\theta$ is the parameter of the model, $\mathcal{E} \in \mathbb{R}^{|\mathcal{B}| \times 1}$is the small perturbation vector and $\mathcal{L}$ is the loss function. 

The goal of Eq. \ref{equ:theta_eb} is to find an optimum parameter $\hat{\theta}_{\mathcal{E},\mathcal{B}}$ so that the loss can be minimized. Then we can use chain rule to compute the impact of perturbation on training batch will pose to the loss:
\begin{equation}
\frac{d\mathcal{L}(D_T,\hat{\theta}_{\mathcal{E},\mathcal{B}})}{d\mathcal{E}}|_{\mathcal{E}=0}=-\nabla_\theta \mathcal{L}(D_T,\hat\theta)H_{\hat\theta}^{-1}\nabla_{\theta}^\mathrm{T} \mathcal{L}(\mathcal{B},\hat\theta),
\end{equation}
where $D_T$ is the training dataset, $H_{\hat\theta}$ is the Hessian matrix and $H_{\hat\theta} = \nabla_{\theta}^2 \mathcal{L}(\mathcal{B},\hat\theta)$.

Since we want to jointly train data from memory buffer and from training set to emphasize informative timestamps, we change the batch $\mathcal{B}$ to be $\mathcal{B} = \mathcal{B}_{memory}\cup\mathcal{B}_{train}$,
thus making Eq. \ref{equ:theta_eb} to be:
\begin{equation}
\hat{\theta}_{\mathcal{E},\mathcal{B}} = argmin \mathcal{L}(\mathcal{B}_{memory}\cup\mathcal{B}_{train},\theta) + \mathcal{E}^\mathrm{T}\mathcal{L}(\mathcal{B}_{memory}\cup\mathcal{B}_{train},\theta).
\end{equation}

\noindent Before the perturbation and impact calculation, we first train model without perturbation for $\mathcal{N}$ epochs \cite{sun2022exploring} as pseudo update. After $\mathcal{N}$ epochs, we add perturbation and construct simulated test sets as the true test set is not available yet. We denote simulated test set, sampled from memory buffer and corresponding to the $\mathcal{B}_{memory}$ as $\mathcal{D}_{memory}$, another corresponding to the $\mathcal{B}_{train}$ as $\mathcal{D}_{train}$ and is sampled from seen training samples. Then we can compute two influence score $I_{memory}$, $I_{train}$ and final merging $I^*$ as:
\begin{equation}
I_{memory} = \frac{d\mathcal{L}(D_{memory},\hat{\theta}_{\mathcal{E},\mathcal{B}})}{d\mathcal{E}}|_{\mathcal{E}=0},
\end{equation}
\begin{equation}
I_{train} = \frac{d\mathcal{L}(D_{train},\hat{\theta}_{\mathcal{E},\mathcal{B}})}{d\mathcal{E}}|_{\mathcal{E}=0},
\end{equation}
\begin{equation}\label{eq15}
I^* = \gamma^* \cdot I_{train} + (1-\gamma^*) \cdot I_{memory},
\end{equation}
where the $\gamma^*$ is computed in a similar manner as \cite{sun2022exploring}
\begin{equation}
\gamma^* = min\left(max\left(\frac{(I_{train}-I_{memory})^\mathrm{T}I_{train}}{||I_{train}-I_{memory}||_2^2},0\right),1\right).
\end{equation}

\noindent We can then list all timestamps' influence scores in descending order and keep top $\mathcal{M}$ stamps with $\mathcal{M}$ being the fixed memory buffer size. The memory buffer was updated each epoch after $\mathcal{N}$ during each time interval. In this way, the informative temporal pattern can be consolidated and emphasized.

\subsection{Relation Importance loss Optimization}
During the training procedure, both loss function and RI score guided parameter updating are conducted to allow model balance between minimizing the loss as well as consolidating learned informative invariant pattern, thus achieving long-term accurate prediction.
We adopt elastic weight consolidation (EWC) \cite{kirkpatrick2017overcoming} for loss function guided parameter updating due to its adaptability to evolving spatial temporal data, which has following loss term:
\begin{equation}
\mathcal{L}_{ewc} = \lambda_{ewc} \sum_i F_i(\Psi_{t}(i)-\Psi_{t-1}(i))^2,
\end{equation}
where $\lambda_{ewc}$ refer to the weight of the EWC smoothing term and $F_i$ is the Fisher information of model $\Psi_{t-1}$'s $i$-th parameter $\theta_i$ and is used for measuring the importance of this parameter to the model with calculation formula as:
\begin{equation}
F_i = \frac{1}{|X^{V_{t-1}}|}\sum_{x \in X^{V_{t-1}}}\frac{\partial\mathcal{L}(\theta_i,x)^2}{\partial\theta_i^2},
\end{equation}
where $\mathcal{L}$ is the loss function. Apart from this ordinary smoothing term, we have our RI smoothing term:
\begin{equation}
\mathcal{L}_{RIS} = \lambda_{RIS} \sum_i F_i^{RIS}(\Psi_{t}(i)-\Psi_{t-1}(i))^2,
\end{equation}
\begin{equation}
F_i^{RIS} = \frac{1}{|X^{V_{t-1}}|}\sum_{x \in X^{V_{t-1}}}\frac{\partial RI(\theta_i,x)^2}{\partial\theta_i^2},
\end{equation}
where $\lambda_{RIS}$ refer to the weight of the RI smoothing term and $F_i^{RIS}$ is the importance of parameter $i$ to the changing of RI value of nodes. Finally we have our final RI loss $\mathcal{L}_{RI}$ as:
\begin{equation}\label{eq21}
\mathcal{L}_{RI} = \mathcal{L} + \mathcal{L}_{ewc} + \mathcal{L}_{RIS}.
\end{equation}
The algorithm of INF-GNN is presented below:
\begin{algorithm}
    \caption{Training procedure for \textbf{INF-GNN}}
    \begin{algorithmic}[1]
        \Statex \textbf{Input:} Spatial temporal data set $\{X^{V_1},...,X^{V_T}\}$, training epochs $\mathcal{I}$
        \Statex \textbf{Output:} Optimum parameter $\hat{\theta}$
        \For{$t = 1,...,T$}
            \For{$v \in V^t$}
                \State Calculate $RI(v)$
            \EndFor
            \State Forming informative subgraph $G_{if}$ as Section~\ref{sec4.2}
            \State Cropping $X^{V_t}$ to $X^{V_{if}}$ according to $G_{if}$
            \State Neighbors merging $v^{\prime} \in V_{t}\setminus V_{t-1}$ and add each $v^{\prime}$ to $G_{if}$, its simulation to $X^{V_{if}}$
            \For{$i = 1,...,\mathcal{I}$}
                \If{$t = 1$} 
                    \State  Initialize memory buffer randomly
                \EndIf
                \State Merging $\mathcal{B}_{memory}$ and $\mathcal{B}_{train}$ to be $\mathcal{B}$
                \If{$i<\mathcal{N}$}
                     \State Pseudo update
                \Else 
                    \State Simulate $\mathcal{D}_{memory}$ and $\mathcal{D}_{train}$ and calculate $I^*$ via Eq.~\ref{eq15}
                    \State Replace the memory buffer as Section~\ref{sec4.3}
                \EndIf
                \State Update model by minimizing RI loss $\mathcal{L}_{RI}$ in Eq.~\ref{eq21}
            \EndFor
        \EndFor
    \end{algorithmic}
\end{algorithm}

\section{Experiments}
\subsection{Dataset}
To demonstrate our method's effectiveness, we use the widely known real-world traffic dataset, PEMS3-Stream \cite{chentrafficstream}, which is recorded by California Transportation Agencies (CalTrans) Performance Measurement System (PeMS). PEMS3-Stream consists of 2011-2017 years' data and we mainly focus on records start from July 10th to August 9th. The reasons for choosing PEMS3-Stream are as follows: (1) This dataset records expanding traffic network under considerable data drift due to its reliable description of urban traffic network in California that underwent rapid development from 2011 to 2017. (2) Traffic information is recorded every 30 seconds and then aggregated to 5 minutes, which ensures the capture of even tiny distribution perturbation. More information about the dataset is shown in Table~\ref{tab1}.

\begin{table}[t]
\caption{{\bfseries PEMS3-Stream} Dataset Statistics.}\label{tab1}
\centering
\setlength{\tabcolsep}{2mm}{
\begin{tabular}{l|l|l|l|l|l|l|l}
\hline
\bfseries Year& \bfseries2011 & \bfseries 2012 & \bfseries 2013 & \bfseries 2014 & \bfseries 2015 & \bfseries 2016 & \bfseries 2017 \\
\hline
\bfseries Nodes& 655 &715&786&822&834&850&871\\
\hline
\bfseries Edges& 1577 &1929&2316&2536&2594&2691&2788\\
\hline
\end{tabular}}
\vspace{-0.5cm}
\end{table}

\subsection{Baselines}
We select following baseline methods for comparison:
\begin{itemize}
    \item[$\bullet$]{GRU \cite{9485098}}: Gated Recurrent Unit (GRU) is a variant of RNN using a gating mechanism. We train a new GRU model with all training data each year.
    \item[$\bullet$]{TrafficStream \cite{chentrafficstream}}: TrafficStream is a continual learning strategy based on Jensen-Shannon divergence only on nodes level. It further uses stable nodes, randomly sampled nodes and newly added nodes to form subgraph.
    \item[$\bullet$]{IGNNK-KNN \cite{wu2021inductive}}: A K-nearest neighbors kriging method that use mean value of K-nearest neighbors of an unknown nodes to simulate its data. The simulated data is then combined with training data to train SurModel.
    \item[$\bullet$]{SurModel}: The surrogate model introduced in Section~\ref{sec4.1} that retrained on all nodes of each year.
    \item[$\bullet$]{SurModel-Retrain}: Surrogate model is retrained on all nodes of each year, the trained model is then used for initialization for model in the next year.
    \item[$\bullet$]{SurModel-Expand}: Surrogate model is retrained only on new nodes each year and is initialized on previous year's model.
    \item[$\bullet$]{INF-GNN}: Our proposed Informative Graph Neural Network (INF-GNN) adapts informative subgraph to simulate new entities. It also utilizes temporal memory buffer with each year's influential timestamps to assist model in emphasizing important temporal patterns. Additionally, RI loss based optimization is designed to consolidate patterns.
\end{itemize}

\subsection{Experimental Settings}
We follow the standard to split the training, validation and testing dataset to 6:2:2 ratio. Baseline methods and our models are first trained on $M$ timestamps data from last year and then tested directly on $K$ timestamps data in next year without additional training. In other words, we will train models on 2011 to 2016 and test their performance on 2012 to 2017 correspondingly. Here we set $M = K = 12$. Adam Optimizer \cite{kingma2014adam} is used for optimization and the learning rate is set to 0.01. The memory buffer size $\mathcal{M}$ is set to be 1000 and the pseudo update epoch $\mathcal{N}$ is set to 45 with total training epoch set to 50 epochs, batch size set to 128 for each year. The number of the nodes in informative subgraph is set to be $10\%$ of the whole graph. As for $\lambda_{RIS}$ and $\lambda_{ewc}$, we assign them with equal half-weight proportion. The simulated test set $\mathcal{D}_{memory}$ and $\mathcal{D}_{train}$'s size are set to 100. Mean Absolute Errors (MAE), Root Mean Squared Errors (RMSE) and Mean Absolute Percentage Errors (MAPE) are utilized as metrics. 

\subsection{Prediction Results}
\begin{figure}[t]
\includegraphics[width=12cm]{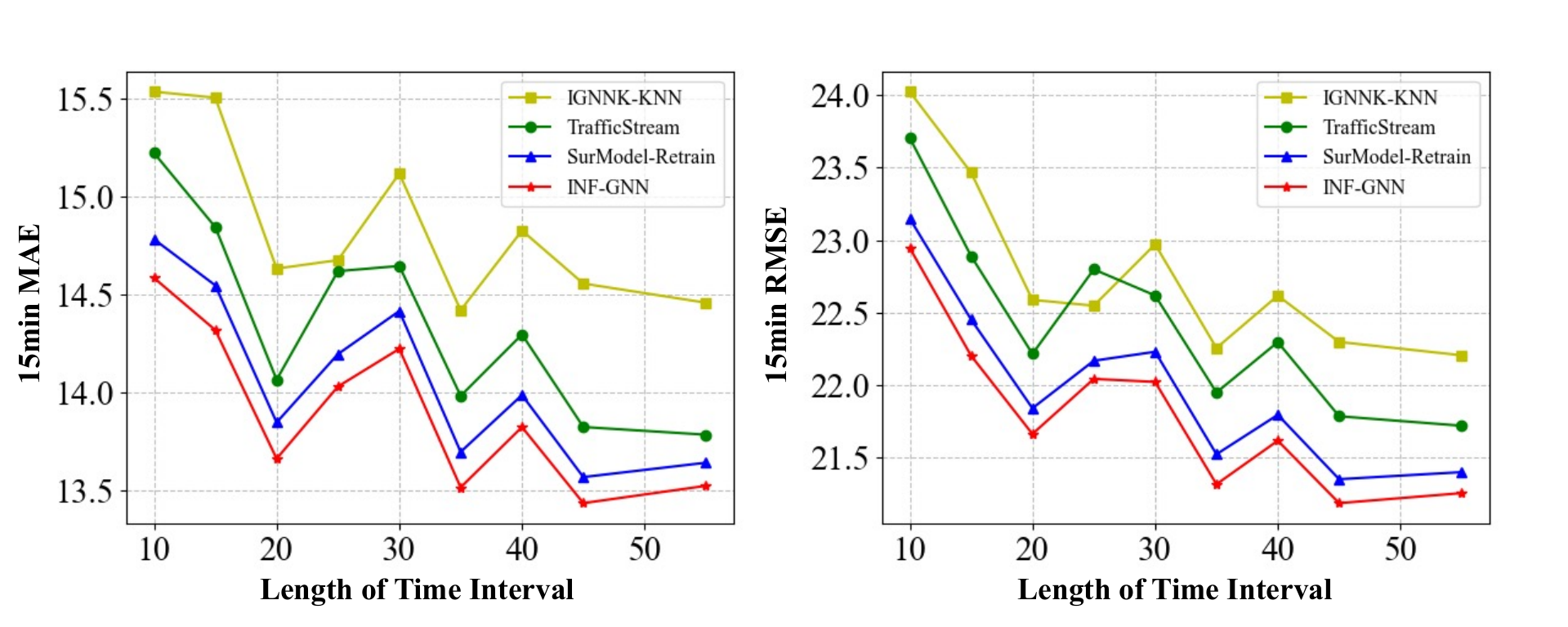}
\caption{Prediction accuracy comparison across different length of time interval } \label{fig3}
\end{figure}

As shown in~\cref{tab2,tab3,tab4}, we present the average MAE, RMSE and MAPE of all models' prediction on existing nodes, new nodes and all nodes. We further vary the length of time interval and the result is shown in Fig.~\ref{fig3}.

\begin{table}[t]
\vspace{-0.3cm}
\caption{Prediction performance of {\bfseries all nodes} on {\bfseries PEMS3-Stream} dataset.}\label{tab2}
\centering
\setlength{\tabcolsep}{1mm}{
\scalebox{0.9}{
\begin{tabular}{l|l|l|l|l|l|l|l|l|l}
\hline
\multirow{2}*{\bfseries Model}& \multicolumn{3}{|l|}{\bfseries15min} & \multicolumn{3}{|l|}{\bfseries30min} & \multicolumn{3}{|l}{\bfseries60min}\\
\cline{2-10}
 &MAE&RMSE&MAPE&MAE&RMSE&MAPE&MAE&RMSE&MAPE\\
\hline
\bfseries IGNNK-KNN& 14.41 &22.21&26.78&15.43&24.21&28.07&17.61&28.13&31.32\\
\bfseries SurModel-Expand& 14.06 &22.16&23.09&15.31&24.47&24.31&17.90&29.03&27.16\\
\bfseries GRU& 13.87 &22.03&24.51&14.78&23.68&25.45&16.87&\bfseries27.28&28.23\\
\bfseries TrafficStream& 13.75 &21.70&21.76&14.89&23.89&23.08&17.20&28.01&26.52\\
\bfseries SurModel& 13.82 &21.71&23.49&14.87&23.80&24.53&17.11&27.82&27.12\\
\bfseries SurModel-Retrain& 13.54 &21.35&23.53&14.62&23.45&24.55&16.88&27.45&27.35\\
\bfseries INF-GNN (DASFAA)& \bfseries13.36 &\bfseries21.18&\bfseries21.62&\bfseries14.50&\bfseries23.32&\bfseries22.85&\bfseries16.83&27.47&\bfseries26.45\\
\bfseries CINF-GNN (ours) & \bfseries13.20 &\bfseries20.99&\bfseries21.50&\bfseries14.38&\bfseries23.17&\bfseries22.80&\bfseries16.77&27.38&\bfseries26.14\\
\hline
\end{tabular}}}
\vspace{-0.2cm}
\end{table}

\begin{table}[t]
\caption{Prediction performance of {\bfseries existing nodes} on {\bfseries PEMS3-Stream} dataset.}\label{tab3}
\centering
\setlength{\tabcolsep}{1mm}{
\scalebox{0.9}{
\begin{tabular}{l|l|l|l|l|l|l|l|l|l}
\hline
\multirow{2}*{\bfseries Model}& \multicolumn{3}{|l|}{\bfseries15min} & \multicolumn{3}{|l|}{\bfseries30min} & \multicolumn{3}{|l}{\bfseries60min}\\
\cline{2-10}
 &MAE&RMSE&MAPE&MAE&RMSE&MAPE&MAE&RMSE&MAPE\\
\hline
\bfseries IGNNK-KNN& 14.50 &22.31&25.72&15.53&24.32&26.98&17.73&28.26&30.12\\
\bfseries SurModel-Expand& 14.14 &22.25&22.18&15.41&24.57&23.37&18.02&29.15&26.13\\
\bfseries GRU& 13.96 &22.13&23.47&14.87&23.79&24.39&16.98&\bfseries27.41&27.10\\
\bfseries TrafficStream& 13.84 &21.80&21.00&14.99&24.00&22.29&17.32&28.14&25.60\\
\bfseries SurModel& 13.90 &21.81&22.55&14.97&23.90&23.59&17.23&27.94&26.14\\
\bfseries SurModel-Retrain& 13.62 &21.44&22.58&14.71&23.56&23.60&16.99&27.57&26.35\\
\bfseries INF-GNN (ours) & \bfseries13.45&\bfseries21.27&\bfseries20.81&\bfseries14.59&\bfseries23.42&\bfseries22.03&\bfseries16.94&27.60&\bfseries25.50\\
\hline
\end{tabular}}}
\vspace{-0.2cm}
\end{table}

\begin{table}
\caption{Prediction performance of {\bfseries new nodes} on {\bfseries PEMS3-Stream} dataset.}\label{tab4}
\centering
\setlength{\tabcolsep}{1mm}{
\scalebox{0.9}{
\begin{tabular}{l|l|l|l|l|l|l|l|l|l}
\hline
\multirow{2}*{\bfseries Model}& \multicolumn{3}{|l|}{\bfseries15min} & \multicolumn{3}{|l|}{\bfseries30min} & \multicolumn{3}{|l}{\bfseries60min}\\
\cline{2-10}
 &MAE&RMSE&MAPE&MAE&RMSE&MAPE&MAE&RMSE&MAPE\\
\hline
\bfseries IGNNK-KNN& 11.75 &18.22&50.22&12.45&19.71&52.10&14.05&22.82&58.17\\
\bfseries SurModel-Expand& 11.68 &18.58&46.75&12.56&20.40&48.49&14.55&24.23&54.03\\
\bfseries GRU& 11.41 &17.88&49.52&12.03&19.16&50.69&13.59&\bfseries22.16&55.28\\
\bfseries TrafficStream& 11.21 &17.78&40.26&11.97&19.40&42.00&13.65&22.58&48.79\\
\bfseries SurModel& 11.27 &17.79&44.12&12.02&19.40&45.80&13.71&22.69&49.78\\
\bfseries SurModel-Retrain& 11.11 &17.54&44.33&11.85&19.15&45.65&13.53&22.36&50.57\\
\bfseries INF-GNN (ours) & \bfseries10.89&\bfseries17.39&\bfseries39.42&\bfseries11.66&\bfseries19.06&\bfseries40.73&\bfseries13.41&22.48&\bfseries47.42\\
\hline
\end{tabular}}}
\end{table}

By analyzing the result, we find that: (1) INF-GNN consistently outperforms other methods across different granularities (15 minutes, 30 minutes, 60 minutes) and shows state-of-art performance. These results show our models can not only counter data drift by maintaining high prediction accuracy on existing nodes, but also generalize well to new nodes by having satisfying performance on new nodes. Furthermore, INF-GNN strikes a balance between learning on existing nodes and new nodes by achieving lowest prediction error on all nodes compared with other baselines. (2) INF-GNN shows stability of generation in different lengths of time intervals, illustrating its advantageous long-term and short-term prediction performance.

\subsection{Ablation Study} In this section, we study how three main components, informative subgraph, informative temporal memory buffer and RI smoothing term, will impact INF-GNN by removing each respectively. Besides, we also study how changing two parameters, weight of RI smoothing term and memory buffer size, will impact INF-GNN by varying each parameter settings separately. 
\begin{figure}[t]
\includegraphics[width=12cm]{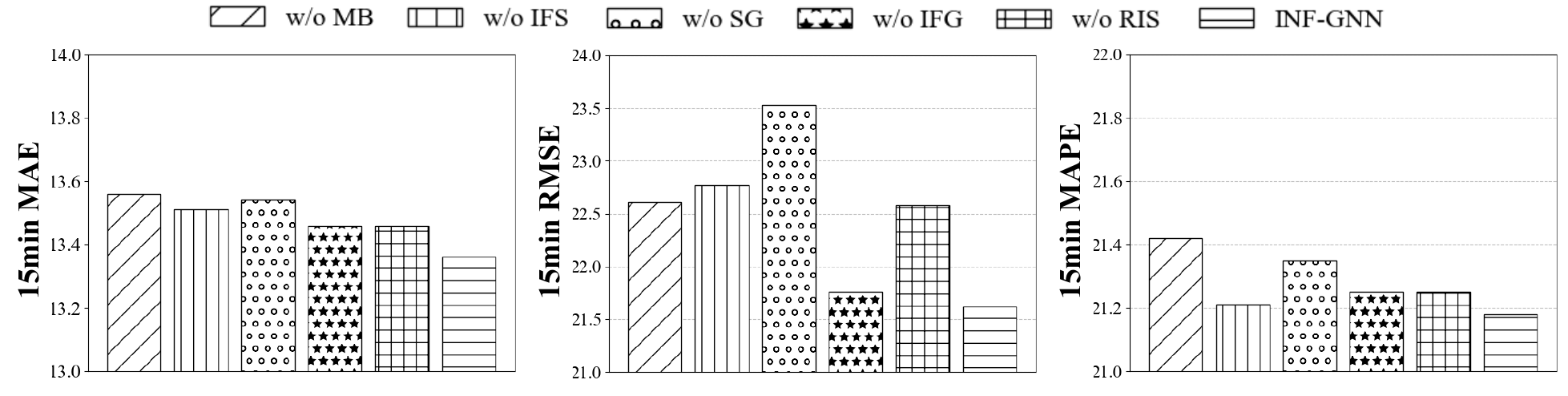}
\caption{Impact of three main components.} \label{fig4}
\vspace{-0.2cm}
\end{figure}

\begin{figure}[t]
\centering
\subfigure{\includegraphics[width=12cm]{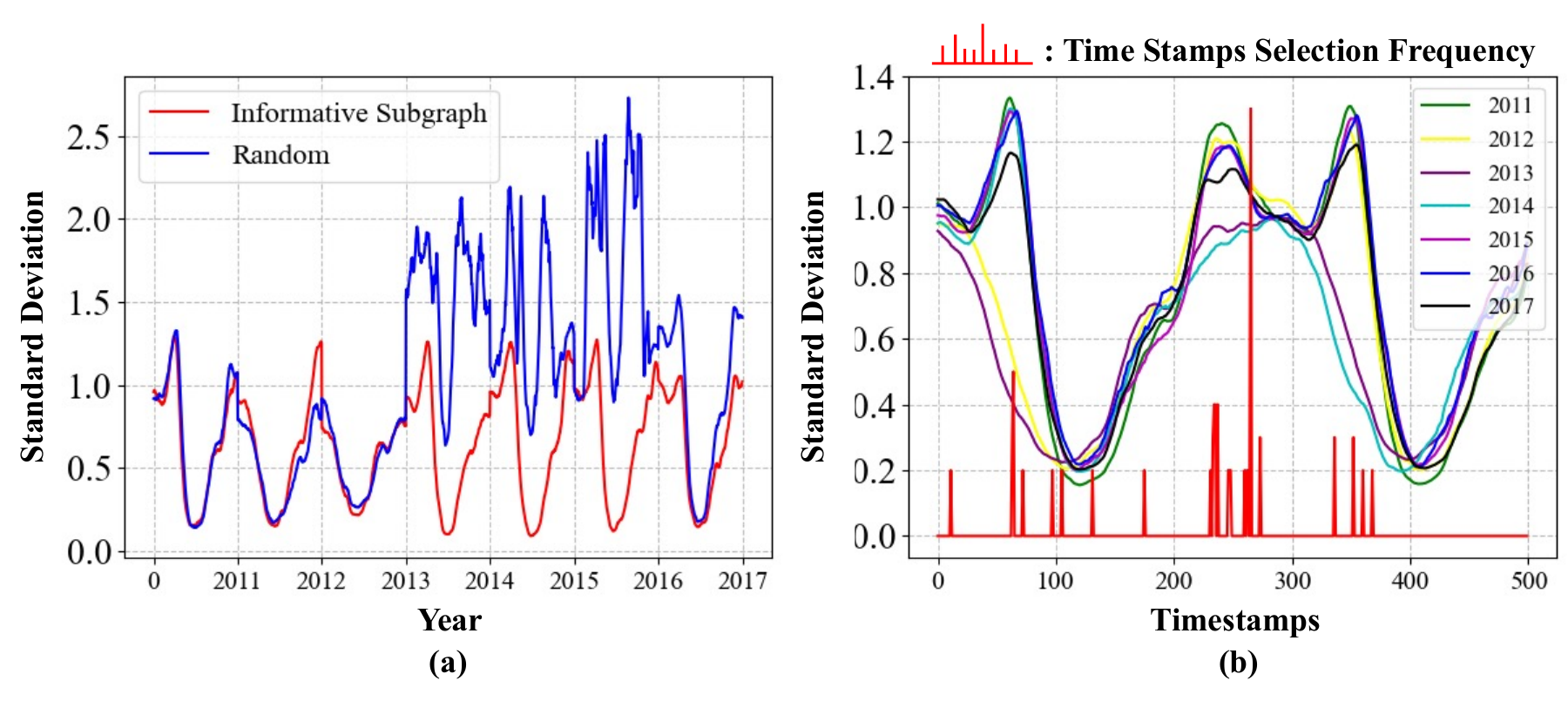}}
\caption{Visualization of components. (a) Informative subgraph contains entities with stable, but high deviation distribution reflecting its invariant and informative characteristic. (b) Red vertical value indicates the frequency of stamps being selected to informative temporal memory buffer. Those with bigger deviation gap will be more frequently selected } \label{fig5}
\vspace{-0.2cm}
\end{figure}

\smallskip\noindent{\bfseries Impact of Informative Subgraph.}
Here we construct: (1) w/o SG: Use whole graph instead of subgraph. (2) w/o IFG: Randomly constructed subgraph. 
From Fig.~\ref{fig4} we can see that using subgraph can achieve better performance compared with using whole graph since subgraph can mitigate negative effects caused by redundant information as well as data drift. Besides, using informative subgraph can distill model with more informative and invariant feature, thus achieving best prediction accuracy. We further visualize informative subgraph selected nodes in Fig.~\ref{fig5}(a) that possess feature deviations that occur repeatedly over time, representing their informative and invariant feature distribution.

\smallskip\noindent{\bfseries Impact of Informative Temporal Memory Buffer.}
Here we construct: (1) w/o MB: Remove memory buffer. (2) w/o IFS: Memory buffer store timestamps randomly. 
From Fig.~\ref{fig4} we can see that removing memory buffer prevents model from emphasizing particular timestamps and randomly storing timestamps will lead to model focusing on timestamps not influential. INF-GNN considers memory buffer with influential timestamps and obtains better performance than w/o IFS and w/o MB. We further visualize stored influential timestamps in Fig.~\ref{fig5}(b). Those with huge deviation gaps between time intervals and encompass more information are more frequently selected.

\smallskip\noindent{\bfseries Impact of RI smoothing term.}
Here we construct (1) w/o RIS: Remove RI smoothing term. From Fig.~\ref{fig4} we can see removing RI smoothing term will lead to the forgetting of learned patterns and prediction accuracy decrease.

\begin{figure}[t]
\includegraphics[width=12cm]{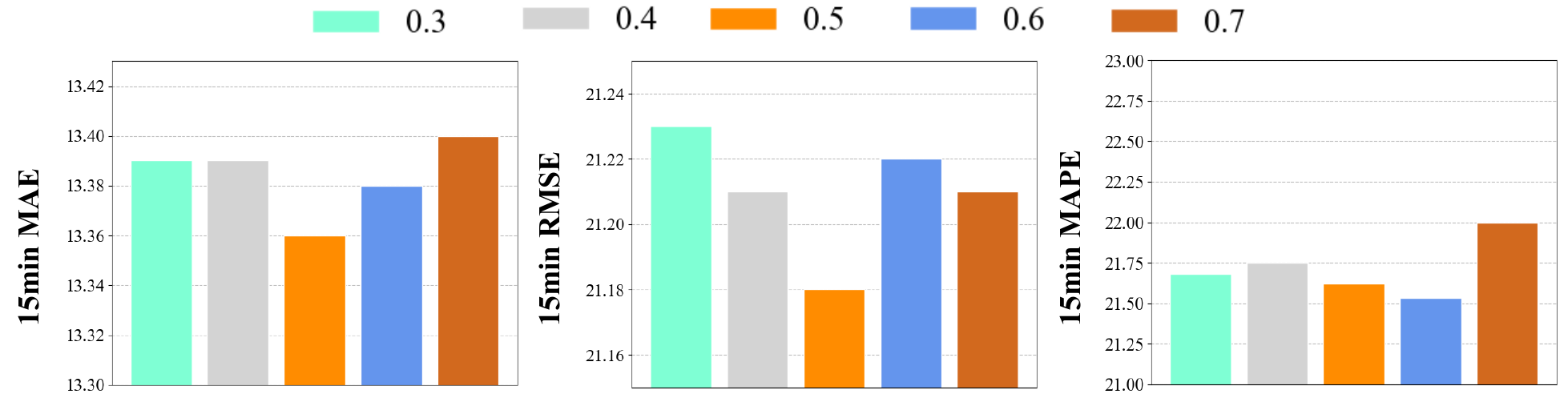}
\caption{Impact of variation on RI smoothing weight.} \label{fig6}
\end{figure}

\smallskip\noindent{\bfseries Impact of variation on RI smoothing weight.} From Fig.~\ref{fig6}, we find 0.5 is the setting to reach balance, since too much emphasis on loss gradient will lose information about informative and invariant spatial temporal patterns while too much emphasis on RI will prevent model from updating parameters according to prediction accuracy.

\begin{figure}[t]
\includegraphics[width=12cm]{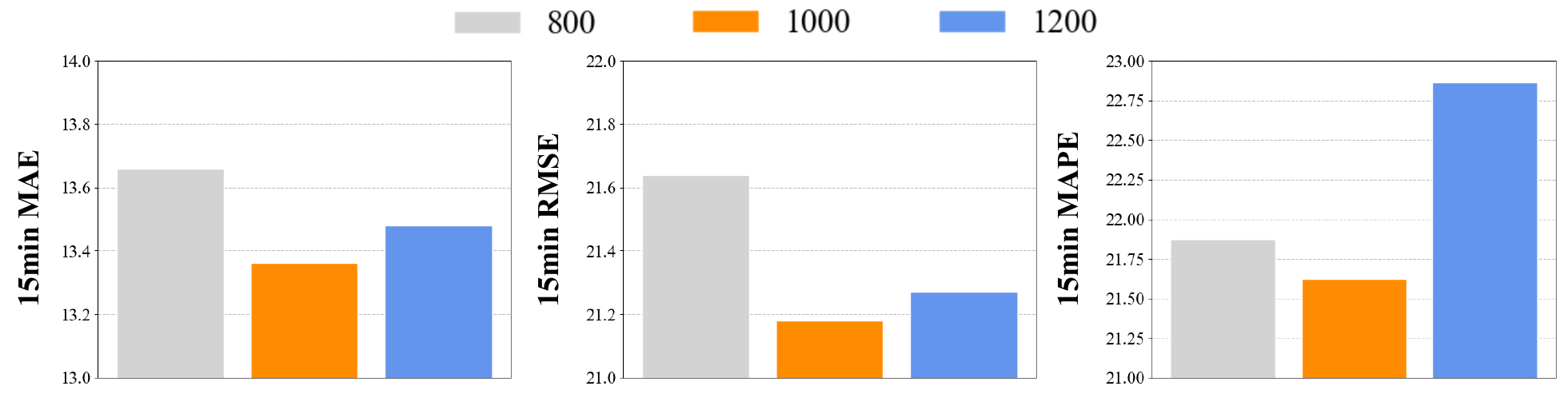}
\caption{Impact of variation on memory buffer size.} \label{fig7}
\vspace{-0.2cm}
\end{figure}

\smallskip\noindent{\bfseries Impact of variation on memory buffer size.} From Fig.~\ref{fig7}, we can observe that as memory buffer size increases from 800 to 1200, the performance will first increase and then decrease, demonstrating the 1000 buffer size to be the best setting. This observation implies that while small memory buffer size will cause the timestamps stored to be replaced too frequently and hurt performance, big memory buffer will lower the frequency of replacement, which prevents model from capturing pattern variation in time.

\section{Conclusion}

In this paper, an Informative Graph Neural Network (INF-GNN) is proposed to perform inductive spatial temporal prediction under data drift. Specifically, an informative subgraph is constructed with invariant entities, which can be utilized for generalizing new entities. Then a memory buffer composed of informative timestamps is constructed to enable INF-GNN emphasize influential timestamps and better capture temporal patterns evolving. Additionally, we design RI loss optimization for pattern consolidation. Experiments on the PEMS3-Stream dataset under severe data drift, further verifying our model show state-of-art prediction accuracy on both existing old nodes as well as new nodes after graph expansion. In the future, we plan to investigate inductive prediction with data drift in other application fields.
~\\

\noindent{\bfseries Acknowledgements.} This work is partially supported by HK RGC Research Impact Fund No.: R5034-18, HK RGC Theme-based Research Scheme No.: T41-603/20-R and the Research Institute for Artificial Intelligence of Things, The Hong Kong Polytechnic University.

%
%
%
%
\bibliographystyle{splncs04}
\bibliography{citation}

\begin{thebibliography}{10}
\providecommand{\url}[1]{\texttt{#1}}
\providecommand{\urlprefix}{URL }
\providecommand{\doi}[1]{https://doi.org/#1}

\bibitem{andreoletti2019network}
Andreoletti, D., Troia, S., Musumeci, F., Giordano, S., Maier, G., Tornatore, M.: Network traffic prediction based on diffusion convolutional recurrent neural networks. In: IEEE INFOCOM 2019-IEEE Conference on Computer Communications Workshops. pp. 246--251. IEEE (2019)

\bibitem{appleby2020kriging}
Appleby, G., Liu, L., Liu, L.P.: Kriging convolutional networks. In: Proceedings of the AAAI Conference on Artificial Intelligence. vol.~34, pp. 3187--3194 (2020)

\bibitem{chentrafficstream}
Chen, X., Wang, J., Xie, K.: Trafficstream: A streaming traffic flow forecasting framework based on graph neural networks and continual learning. In: Proceedings of the Thirtieth International Joint Conference on Artificial Intelligence, {IJCAI-21}. pp. 3620--3626 (2021)

\bibitem{cressie2015statistics}
Cressie, N., Wikle, C.K.: Statistics for spatio-temporal data. John Wiley \& Sons (2015)

\bibitem{cui2019traffic}
Cui, Z., Henrickson, K., Ke, R., Wang, Y.: Traffic graph convolutional recurrent neural network: A deep learning framework for network-scale traffic learning and forecasting. IEEE Transactions on Intelligent Transportation Systems  \textbf{21}(11),  4883--4894 (2019)

\bibitem{deng2021graph}
Deng, L., Liu, X.Y., Zheng, H., Feng, X., Chen, Y.: Graph spectral regularized tensor completion for traffic data imputation. IEEE Transactions on Intelligent Transportation Systems  \textbf{23}(8),  10996--11010 (2021)

\bibitem{du2021adarnn}
Du, Y., Wang, J., Feng, W., Pan, S., Qin, T., Xu, R., Wang, C.: Adarnn: Adaptive learning and forecasting of time series. In: Proceedings of the 30th ACM international conference on information \& knowledge management. pp. 402--411 (2021)

\bibitem{jin2021community}
Jin, T., Wu, Q., Ou, X., Yu, J.: Community detection and co-author recommendation in co-author networks. International Journal of Machine Learning and Cybernetics  \textbf{12},  597--609 (2021)

\bibitem{kim2021reversible}
Kim, T., Kim, J., Tae, Y., Park, C., Choi, J.H., Choo, J.: Reversible instance normalization for accurate time-series forecasting against distribution shift. In: International Conference on Learning Representations (2021)

\bibitem{kingma2014adam}
Kingma, D.P., Ba, J.: Adam: A method for stochastic optimization. arXiv preprint arXiv:1412.6980  (2014)

\bibitem{kirkpatrick2017overcoming}
Kirkpatrick, J., Pascanu, R., Rabinowitz, N., Veness, J., Desjardins, G., Rusu, A.A., Milan, K., Quan, J., Ramalho, T., Grabska-Barwinska, A., et~al.: Overcoming catastrophic forgetting in neural networks. Proceedings of the national academy of sciences  \textbf{114}(13),  3521--3526 (2017)

\bibitem{koh2017understanding}
Koh, P.W., Liang, P.: Understanding black-box predictions via influence functions. In: International conference on machine learning. pp. 1885--1894. PMLR (2017)

\bibitem{liu2021coinet}
Liu, J., Guo, X., Li, B., Yuan, Y.: Coinet: Adaptive segmentation with co-interactive network for autonomous driving. In: 2021 IEEE/RSJ International Conference on Intelligent Robots and Systems (IROS). pp. 4800--4806. IEEE (2021)

\bibitem{liu2021graph}
Liu, J., Guo, X., Yuan, Y.: Graph-based surgical instrument adaptive segmentation via domain-common knowledge. IEEE Transactions on Medical Imaging  \textbf{41}(3),  715--726 (2021)

\bibitem{liu2021prototypical}
Liu, J., Guo, X., Yuan, Y.: Prototypical interaction graph for unsupervised domain adaptation in surgical instrument segmentation. In: International Conference on Medical Image Computing and Computer-Assisted Intervention. pp. 272--281. Springer (2021)

\bibitem{pareja2020evolvegcn}
Pareja, A., Domeniconi, G., Chen, J., Ma, T., Suzumura, T., Kanezashi, H., Kaler, T., Schardl, T., Leiserson, C.: Evolvegcn: Evolving graph convolutional networks for dynamic graphs. In: Proceedings of the AAAI conference on artificial intelligence. vol.~34, pp. 5363--5370 (2020)

\bibitem{passalis2019deep}
Passalis, N., Tefas, A., Kanniainen, J., Gabbouj, M., Iosifidis, A.: Deep adaptive input normalization for time series forecasting. IEEE transactions on neural networks and learning systems  \textbf{31}(9),  3760--3765 (2019)

\bibitem{pearl2016causal}
Pearl, J., Glymour, M., Jewell, N.P.: Causal inference in statistics: A primer. John Wiley \& Sons (2016)

\bibitem{9485098}
Shu, W., Cai, K., Xiong, N.N.: A short-term traffic flow prediction model based on an improved gate recurrent unit neural network. IEEE Transactions on Intelligent Transportation Systems  \textbf{23}(9),  16654--16665 (2022)

\bibitem{sun2022exploring}
Sun, Q., Lyu, F., Shang, F., Feng, W., Wan, L.: Exploring example influence in continual learning. Advances in Neural Information Processing Systems  \textbf{35},  27075--27086 (2022)

\bibitem{tang2020joint}
Tang, X., Yao, H., Sun, Y., Aggarwal, C., Mitra, P., Wang, S.: Joint modeling of local and global temporal dynamics for multivariate time series forecasting with missing values. In: Proceedings of the AAAI Conference on Artificial Intelligence. vol.~34, pp. 5956--5963 (2020)

\bibitem{wang2023pattern}
Wang, B., Zhang, Y., Wang, X., Wang, P., Zhou, Z., Bai, L., Wang, Y.: Pattern expansion and consolidation on evolving graphs for continual traffic prediction. In: Proceedings of the 29th ACM SIGKDD Conference on Knowledge Discovery and Data Mining. pp. 2223--2232 (2023)

\bibitem{wang2020streaming}
Wang, J., Song, G., Wu, Y., Wang, L.: Streaming graph neural networks via continual learning. In: Proceedings of the 29th ACM international conference on information \& knowledge management. pp. 1515--1524 (2020)

\bibitem{wang2022causal}
Wang, W., Lin, X., Feng, F., He, X., Lin, M., Chua, T.S.: Causal representation learning for out-of-distribution recommendation. In: Proceedings of the ACM Web Conference 2022. pp. 3562--3571 (2022)

\bibitem{wu2021inductive}
Wu, Y., Zhuang, D., Labbe, A., Sun, L.: Inductive graph neural networks for spatiotemporal kriging. In: Proceedings of the AAAI Conference on Artificial Intelligence. vol.~35, pp. 4478--4485 (2021)

\bibitem{xu2020ge}
Xu, D., Wei, C., Peng, P., Xuan, Q., Guo, H.: Ge-gan: A novel deep learning framework for road traffic state estimation. Transportation Research Part C: Emerging Technologies  \textbf{117},  102635 (2020)

\bibitem{yang2023higrn}
Yang, H., Li, W., Hou, S., Guan, J., Zhou, S.: Higrn: A hierarchical graph recurrent network for global sea surface temperature prediction. ACM Transactions on Intelligent Systems and Technology  (2023)

\bibitem{zhang2022dynamic}
Zhang, Z., Wang, X., Zhang, Z., Li, H., Qin, Z., Zhu, W.: Dynamic graph neural networks under spatio-temporal distribution shift. Advances in Neural Information Processing Systems  \textbf{35},  6074--6089 (2022)

\bibitem{zheng2023increase}
Zheng, C., Fan, X., Wang, C., Qi, J., Chen, C., Chen, L.: Increase: Inductive graph representation learning for spatio-temporal kriging. In: Proceedings of the ACM Web Conference 2023. pp. 673--683 (2023)

\end{thebibliography}

\end{document}